\documentclass[twocolumn]{article}

\usepackage[utf8]{inputenc}
\usepackage[margin=1.25in]{geometry}
\usepackage{authblk}
\usepackage{setspace}
\usepackage{graphicx}
\graphicspath{{./figures/}}
\usepackage{subcaption}
\usepackage{amsmath}
\usepackage{amssymb}
\usepackage{booktabs}
\usepackage{hyperref}
\usepackage{soul}

\usepackage[switch]{lineno}       
\usepackage[
  style=nejm,
  citestyle=numeric-comp,
  sorting=none,
  maxbibnames=99
]{biblatex}
\title{RA-CMF: Region-Adaptive Conditional MeanFlow for CT Image Reconstruction}

\author[1*]{Md Shifatul Ahsan Apurba}
\author[2]{Md Selim}
\author[1]{Jin Chen}

\affil[1]{Department of Biomedical Informatics and Data Science, University of Alabama at Birmingham, Birmingham, Alabama, United States.}
\affil[2]{Department of Computer Science, Florida Polytechnic University, Lakeland, Florida, United States.}
\affil[*]{Address correspondence to: mapurba@uab.edu}

\date{}

\begin{document}
\twocolumn[
\begin{@twocolumnfalse}

\maketitle

\begin{abstract}
The use of CT imaging is important for screening, diagnosis, therapy planning, and prognosis of lung cancers. Unfortunately, due to differences in imaging protocols and scanner models, CT images acquired by different means may show large differences in noise statistics, contrast, and texture. In this study, we develop a novel conditional MeanFlow pipeline for CT image reconstruction. We introduce a conditional MeanFlow network that models the reconstruction trajectory by predicting image-conditioned flow fields given intermediate image states. The image reconstruction network is trained with a MeanFlow consistency loss along with the image reconstruction loss. In order to provide a spatially adaptive refinement process, we integrate a regional reinforcement learning-driven policy network into our approach. The policy network receives information about the MeanFlow rollouts and provides predictions in terms of tile-wise refinement budgets, stopping criteria, and total budget allocation of refinement processes. Our policy network is trained through reinforcement learning in a policy gradient framework, where the goal of the training reward is to maximize reconstruction quality while minimizing unnecessary computations and avoiding instabilities. In this way, our approach combines conditional flow-based reconstruction with reinforcement learning–based spatial reconstruction control. This allows our approach to allocate additional refinement to difficult regions while stabilizing areas already showing sufficient quality. Experimentally, we validate the effectiveness of our approach for CT image reconstruction. Our results show high accuracy in the tumor ROI, with the average radiomic feature CCC being 0.93 $\pm$ 0.09, an average PSNR of 31.94 $\pm$ 2.64, and average SSIM of 0.97 $\pm$ 0.03. Moreover, there is an improvement in the overall quality of images, with an average PSNR of 34.23 $\pm$ 1.71 and average SSIM of 0.95 $\pm$ 0.01.
\end{abstract}
\vspace{1em}
\end{@twocolumnfalse}
]


\section{Introduction}
Lung cancer is the primary cause of cancer-related deaths and continues to be the most common form of cancer around the globe \cite{siegel2024cancer,who2026lungcancer}. For instance, it was estimated that lung and bronchus cancer would be associated with 226,650 incident cases and 124,730 deaths in 2025 in the United States \cite{siegel2025cancer}. Overall, the 5-year survival rate of lung cancer is 28.1\% but improves up to 64.7\% upon detection of lung cancer at the localized stage \cite{seer2026lungbronchus}. CT is instrumental for lung cancer screening, diagnosis, therapeutic planning, and prognosis. Specifically, lung cancer mortality can be reduced by roughly 20\% via low-dose CT screening \cite{aberle2011reduced}. Besides detection, CT imaging allows for extracting quantitative features reflecting tumor phenotypes, heterogeneity, and dynamics to facilitate clinical decisions \cite{aerts2014decoding,lambin2017radiomics}.

However, despite the high value of CT imaging in the clinical setting, there are many sources of variability affecting CT images, which may include scanner vendors, imaging protocols, and reconstruction settings. Such variability results in different properties of CT images, including their noise level, contrast, and texture. The inconsistency can appear in both the same scanner using various imaging protocols and different scanners from various vendors \cite{mackin2015measuring}. Notably, the impact of imaging protocol inconsistency on CT image quality is not limited to the visual appearance of images, since it can negatively affect quantitative feature extraction from CT images. Moreover, variability introduced by imaging protocols can equalize or even surpass biological differences between tumors \cite{mackin2015measuring}. Such variability poses serious barriers for conducting large-scale multi-center experiments and adversely affects the robustness of lung cancer analysis based on quantitative CT imaging.

Various post-image reconstruction approaches have been proposed as part of image harmonization tasks, typically focusing on specific anatomical regions and ineffective in clinical practice \cite{orlhac2018postreconstruction}. Image-level techniques involving deep learning showed promising performance; for example, a series of CT images across different reconstruction kernels can be transformed to increase the reproducibility of radiomic features \cite{choe2019deep}. Recently, diffusion-based models, such as DiffusionCT \cite{selim2024diffusionct} proved to be effective in capturing complex distributions of images for CT standardization purposes. However, this approach depends on the careful selection of diffusion steps, and the results can become unstable across the z-dimension, leading to poor 3D feature consistency. Also, these techniques still rely on globally uniform transformation and ignore image degradation heterogeneity. MeanFlow has recently been introduced as an efficient generative modeling framework that learns average velocity fields over finite time intervals, enabling high-quality one-step or few-step generation \cite{geng2025meanflows}.

In this work, we propose a region-adaptive conditional MeanFlow framework for image-domain CT reconstruction that combines image-conditioned flow-based reconstruction with region-aware adaptive refinement. The proposed framework is designed to selectively allocate refinement to spatially heterogeneous regions while preserving anatomical consistency in areas that already exhibit satisfactory image quality. By integrating conditional MeanFlow modeling with adaptive regional control, our approach aims to improve CT image quality and enhance the reliability of downstream quantitative radiomic analysis.

\section{Background}

\subsection{Radiomic Features}

The radiomic features enable quantification through the analysis of intensity distributions and textures present in images beyond human observation. Radiomic features may capture biological and phenotypic attributes that relate to tissue heterogeneity and biological processes; they can therefore be utilized for quantitative tumor characterization and outcome prediction \cite{lambin2012radiomics,aerts2014decoding}.

Here, radiomic features are extracted based on the PyRadiomics pipeline, a standardized and reliable implementation of radiomic feature families commonly used in research \cite{vangriethuysen2017pyradiomics}. Radiomic features defined here are consistent with the standards of radiomic features proposed by, e.g., the Image Biomarker Standardization Initiative (IBSI) \cite{zwanenburg2020ibsi}. The selected radiomic features are grouped into six categories. The first-order features capture information about voxel intensity characteristics like mean, variance, and entropy, ignoring any spatial interaction between voxels. The GLCM features quantify second-order spatial interactions using the model of gray-level value co-occurrence \cite{haralick1973textural}. The GLDM features measure gray-level dependency relationships between neighboring voxels \cite{sun1983neighboring}. The GLRLM features represent the length of gray-level run sequences \cite{galloway1975texture}. The GLSZM features capture the distribution of sizes of homogeneous gray-level regions regardless of their orientations \cite{thibault2009texture}. The NGTDM features analyze gray-level differences in terms of a given voxel and the gray-levels of voxels in its vicinity. This feature type captures certain perceptual aspects of image texture like contrast, coarseness, and busyness \cite{amadasun1989textural}.

Thus, the radiomic features listed above represent a set of measures of intensity and texture heterogeneity of CT scans, widely adopted by research and used in numerous radiomics studies \cite{lambin2012radiomics,zwanenburg2020ibsi}. However, the radiomic feature values may differ depending on image acquisition, image reconstruction, preprocessing, and discretization parameters. Previous studies have revealed that radiomic features extracted from CT data are influenced both by intra-scanner variation and inter-scanner variation \cite{mackin2015measuring}. Other research showed the impact of voxel size and gray-level normalization on the values of CT radiomic features in lung cancer imaging \cite{shafiq2018voxel}. Notably, sometimes the effect of scanner-specific variability may be equivalent to the effect of biology of different tumors, which poses a great barrier to multi-center radiomics studies and limits the robustness of radiomics-based clinical models \cite{mackin2015measuring,aerts2014decoding}.

\begin{figure}[t]
\centering
\includegraphics[width=\columnwidth]{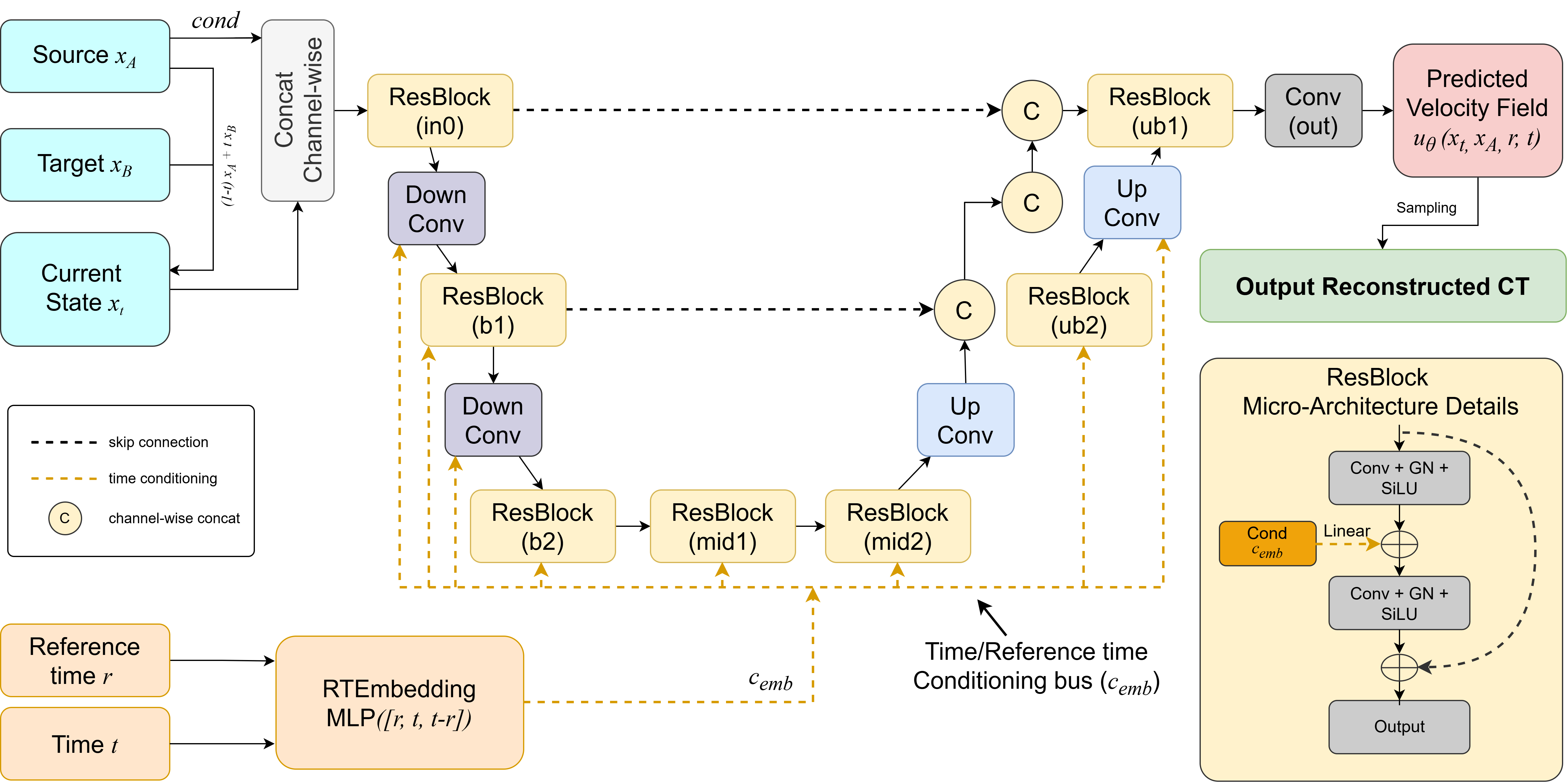}
\caption{Conditional MeanFlow (CMF) backbone architecture. The current state $x_t$ and conditioning image $x_A$ are concatenated channel-wise and processed by a U-Net. The reference time $r$ and time $t$ are encoded through an RT-embedding module and injected into residual blocks to predict the average velocity field $u_\theta(x_t,x_A,r,t)$.}
\label{fig:cmf_arch}
\end{figure}

\begin{figure*}[t]
\centering
\includegraphics[width=0.98\textwidth]{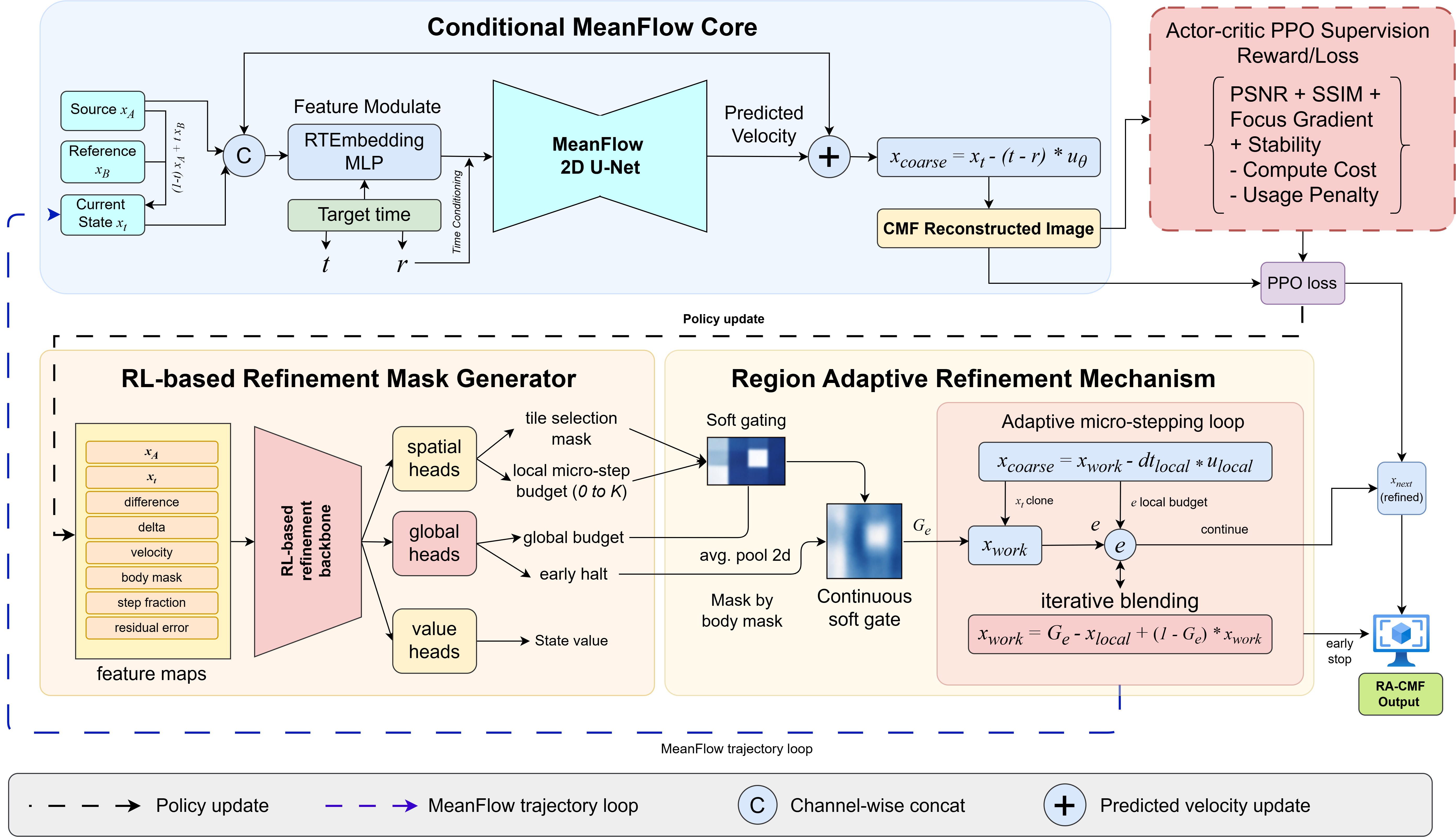}
\captionof{figure}{Region adaptive conditional MeanFlow (RA-CMF) pipeline. The CMF backbone first predicts a coarse refinement update. An actor-critic controller then extracts state features from the current image state, source condition, residual error, velocity magnitude, body mask, and step fraction. The agent predicts tile selection masks, local refinement budgets, a global budget, and an early stopping decision. Selected regions are refined through adaptive local micro-steps using soft spatial gates, while PPO supervision optimizes refinement quality, stability, and computational cost.}
\label{fig:racmf_arch}
\end{figure*}

\subsection{Prior CT Image Reconstruction and Harmonization}

It is difficult to standardize CT acquisition and reconstruction protocols across institutions, primarily due to variations in scanner hardware, clinical requirements, and site-specific workflow constraints. Consequently, post-acquisition harmonization has emerged as a practical strategy for improving consistency in multi-center imaging studies. Feature-level techniques, such as ComBat, have been shown to reduce non-biological variability in extracted imaging biomarkers across different centers \cite{orlhac2022combat}. However, because these approaches are applied after feature extraction, they are inherently limited in their ability to correct spatially localized inconsistencies present in the original CT images.

To address these limitations, image-level deep learning approaches have been investigated to directly transform CT images prior to feature extraction. For instance, a previous study demonstrated that deep learning–based reconstruction kernel conversion can enhance the reproducibility of radiomic features in pulmonary nodules and masses \cite{choe2019deep}. Building on this idea, GAN-based standardization frameworks such as STAN-CT \cite{selim2020stanct} and RadiomicGAN \cite{selim2021radiomicgan} have shown that image-level harmonization can effectively reduce protocol-induced variability and improve radiomics reproducibility. Despite these advantages, adversarial learning approaches are known to suffer from issues such as training instability and mode collapse, which can limit their reliability in quantitative medical imaging applications \cite{arjovsky2017wasserstein}.

More recently, diffusion-based generative models have emerged as a promising alternative for image synthesis and transformation, largely due to their improved stability and modeling capacity. Denoising diffusion probabilistic models introduced a progressive denoising process that enables the generation of high-quality images \cite{ho2020ddpm}. Building on this idea, latent diffusion models reformulate the process in a latent space, making high-resolution image generation more computationally efficient \cite{rombach2022latent}. In the context of CT imaging, DiffusionCT demonstrated that such models can be applied to standardize CT images while maintaining structural fidelity \cite{selim2024diffusionct}. Despite these advances, most existing image-level harmonization approaches still rely on globally applied transformations and do not explicitly address the spatially varying nature of CT image degradation.

In lung CT imaging, degradation is often spatially non-uniform. Regions containing nodules, vessel boundaries, and complex parenchymal structures typically require more aggressive refinement compared to relatively homogeneous regions. This observation motivates the development of enhancement frameworks that can adapt to local image characteristics while maintaining anatomical consistency. Such adaptive strategies are essential for improving the reliability of radiomic features and enabling robust quantitative analysis across diverse imaging settings.

\section{Materials and Methods}

Our work introduces a unified conditional MeanFlow pipeline for CT image reconstruction. Given an input CT image \(x_A \in \mathbb{R}^{H \times W}\) reconstructed using the non-standard kernel, the goal is to reconstruct the corresponding target image \(x_B \in \mathbb{R}^{H \times W}\) represented by the standard reconstruction kernel. In this study, \(x_A\) corresponds to the GE Standard reconstruction and \(x_B\) corresponds to the GE Lung reconstruction obtained from the same CT acquisition. The proposed task operates in the image domain and does not involve reconstruction from raw projection or sparse-view data. The proposed framework consists of two main components: a conditional MeanFlow backbone that models the reconstruction trajectory, and a region-aware controller that adaptively distributes refinement across spatial regions. The proposed region-aware conditional MeanFlow (RA-CMF) architecture is illustrated in Fig.~\ref{fig:racmf_arch}. The MeanFlow backbone is motivated by recent developments in MeanFlow generative modeling, which focuses on learning average velocity fields over finite time intervals rather than only instantaneous velocities, thereby enabling efficient one-step or few-step generation \cite{geng2025meanflows}. In contrast to the original MeanFlow formulation—primarily explored in the context of natural image generation with class-conditional settings; our approach conditions directly on the degraded CT image to perform image-to-image reconstruction. To the best of our knowledge, MeanFlow has not yet been applied to image-domain CT reconstruction, and this work represents an initial step toward exploring conditional MeanFlow for medical image restoration.

The design of the proposed framework is motivated by the observation that the degradation in CT images is often spatially heterogeneous. As a result, applying a uniform reconstruction strategy across the entire image may not be equally effective for all regions. This consideration is particularly relevant in lung CT imaging, where variations in image quality and texture can arise from differences in anatomical structures, acquisition protocols, and reconstruction settings, ultimately influencing the reliability of radiomic features \cite{mackin2015measuring,choe2019deep}. To address this, the region-aware controller is designed to selectively allocate refinement to more challenging regions, while avoiding unnecessary modifications in areas that already exhibit acceptable image quality.

\subsection{Conditional MeanFlow Backbone}

The reconstruction process is formulated as a continuous transformation defined over a normalized time variable $t \in [0,1]$. Rather than learning a direct one-step mapping from $x_A$ to $x_B$, the model is trained using intermediate states constructed between the clean target image and Gaussian noise. Specifically, for a sampled time $t$, we define
\begin{equation}
x_t = (1-t)x_B + te, \qquad e \sim \mathcal{N}(0,I),
\end{equation}
where $x_t$ represents an intermediate point along the trajectory from the target image $x_B$ to a noise realization $e$. This formulation allows the model to learn reconstruction as a progressive transformation process instead of a single deterministic mapping.

An important characteristic of the proposed approach is that the transformation is \emph{conditioned on the source CT image} $x_A$. The network predicts a MeanFlow field
\begin{equation}
u_\theta(x_t, x_A, r, t),
\end{equation}
where $x_t$ denotes the current intermediate state, $x_A$ is the input image, and $(r,t)$ defines a finite time interval with $0 \le r \le t \le 1$. As shown in Fig.~\ref{fig:cmf_arch}, the conditional MeanFlow (CMF) architecture concatenates $x_t$ and $x_A$ along the channel dimension and processes them using a U-Net backbone. The temporal variables $(r,t,t-r)$ are embedded through an RTEmbedding MLP to produce $c_{\mathrm{emb}}$, which is injected into residual blocks throughout the network. This design enables the model to adapt its behavior across different stages of the transformation while preserving the anatomical structures present in $x_A$.

In contrast to standard denoising approaches that estimate instantaneous residuals, the proposed method follows the MeanFlow formulation and predicts the average transformation over a finite time interval. Let
\begin{equation}
v = e - x_B
\end{equation}
denote the direction from the target image to the noise sample. The corresponding MeanFlow target is defined as
\begin{equation}
u^{} = v - (t-r)\frac{d u_\theta}{dt},
\end{equation}
where the time derivative is computed using Jacobian-vector products. The MeanFlow loss is then given by
\begin{equation}
\mathcal{L}_{\mathrm{mf}} =
\mathbb{E}_{x_A,x_B,e,r,t}
\left[
|u_\theta(x_t,x_A,r,t)-u^{}|_1
\right].
\end{equation}

To further encourage accurate reconstruction, the target image is recovered by integrating the predicted transformation from time $t$ back to $0$:
\begin{equation}
\hat{x}0 = x_t - t,u\theta(x_t,x_A,0,t).
\end{equation}
The reconstructed image is supervised using an image-level loss
\begin{equation}
\mathcal{L}_{\mathrm{img}} = |\hat{x}_{0} - x_B|_{1}.
\end{equation}
The overall training objective for the backbone is therefore defined as
\begin{equation}
\mathcal{L}_{\mathrm{base}} =
\mathcal{L}_{\mathrm{mf}} + \lambda_1 \mathcal{L}_{\mathrm{img}}.
\end{equation}

\subsection{MeanFlow Based Adaptive Reconstruction}

After training, the backbone is employed to generate reconstructed images by applying the learned transformation conditioned on $x_A$. In the RA-CMF rollout illustrated in Fig.~\ref{fig:racmf_arch}, the current state of the process is denoted by $x_k$, which corresponds to the state $x_t$ in the architectural formulation. The reconstruction proceeds in a progressive manner, where at each step an initial coarse update is computed as
\begin{equation}
x^{\mathrm{coarse}}_{k+1} = x_k - (t_k - t_{k+1})\,u_\theta(x_k, x_A, t_{k+1}, t_k),
\end{equation}
with $k$ indexing the refinement step.


Rather than applying a uniform update across the entire image, the proposed framework incorporates a region-aware adaptive refinement strategy. Based on the current state, the controller predicts tile-wise refinement levels, spatial masks, a global refinement budget, and a stopping signal. Unlike fixed or heuristic refinement strategies, RA-CMF learns refinement decisions within the conditional MeanFlow trajectory. The controller uses the current image state, source image, residual error, velocity information, and refinement progress to determine where additional updates are needed and when the refinement process should stop. As shown in Fig.~\ref{fig:racmf_arch}, the controller takes as input multiple state-dependent feature maps, including the source image, current state, difference map, update direction, body mask, step fraction, and residual error. These inputs are processed through spatial, global, and value heads to produce refinement actions along with PPO-based supervision.

In our implementation, each CT image is processed at a resolution of
$512 \times 512$ pixels and divided into a $16 \times 16$ spatial grid,
resulting in 256 tiles with an effective size of $32 \times 32$ pixels
per tile. The rollout allows up to 8 coarse MeanFlow steps, while the
controller can allocate up to 16 additional local refinement steps to
selected regions. The global refinement budget is constrained to the
range $[0.05, 0.25]$, and early stopping is determined using a stopping
threshold of 0.95.

The tile-level predictions are translated into spatial masks $M_k^{(m)} \in [0,1]^{H \times W}$, which specify where additional refinement should be applied. These masks act as continuous soft gates, consistent with the region-adaptive refinement mechanism depicted in Fig.~\ref{fig:racmf_arch}. For each refinement level $m$, an additional MeanFlow update is computed and selectively applied within the masked regions:
\begin{equation}
\begin{aligned}
x_{k+1}^{(m)}
={}&
M_k^{(m)} \odot
\Big[
x_{k+1}^{(m-1)}
-\Delta t_{\mathrm{local}} \\
&\qquad \times
u_\theta\!\left(
x_{k+1}^{(m-1)},x_A,t_{k+1},t_k
\right)
\Big] \\
&+
\left(1-M_k^{(m)}\right)
\odot x_{k+1}^{(m-1)} .
\end{aligned}
\label{eq:local_refinement}
\end{equation}
where $\odot$ denotes element-wise multiplication. This formulation ensures that refinement is concentrated in regions that require further improvement, while preserving regions that are already stable.

After applying the selected local refinements, the resulting state defines the updated image $x_{k+1}$. In addition, the controller predicts a global stopping decision, allowing the refinement process to terminate early once sufficient image quality is achieved. As a result, the total number of refinement steps is determined adaptively rather than being fixed in advance.

\subsection{Policy Optimization}

The controller is trained using a reinforcement learning framework, in which an agent learns refinement strategies by maximizing a reward signal obtained through interaction with the reconstruction process \cite{sutton2018reinforcement}. Each image state is treated as the environment state, and the corresponding refinement decisions define the action. In practice, the action space includes tile-selection decisions, local micro-step budgets, a global refinement budget, and an early stopping signal, as illustrated in Fig.~\ref{fig:racmf_arch}. After each refinement step, a reward is computed based on the improvement in image quality:
\begin{equation}
r_k =
\mathcal{Q}(x_{k+1}, x_B) - \mathcal{Q}(x_k, x_B)
- \alpha \|a_k\|_1,
\end{equation}
where $\mathcal{Q}$ evaluates similarity to the target image using image-quality metrics such as PSNR and SSIM \cite{huynhthu2008psnr,wang2004ssim}. Additionally, a focus-gradient term is incorporated to promote sharper local structures, following classical gradient-based focus measures \cite{nayar1989shape}. The penalty term discourages excessive computation and unnecessary refinement. 

For policy optimization, the PSNR and SSIM reward terms are weighted by
0.02 and 0.25, respectively. The focus-image and focus-gradient terms
use weights of 4.0 and 1.0, while stability penalties of 0.20 and 0.05
are applied to the image and gradient terms, respectively. The
computation penalty is increased from 0.01 to 0.05 during training, and
a stopping bonus of 0.10 is used for appropriate early termination.
PPO uses a clipping parameter of 0.20, a value-loss coefficient of
0.50, an entropy coefficient of 0.01, and 4 optimization epochs per
update. The stopping teacher target uses a margin of 0.02.

The policy is optimized using a PPO-based actor-critic framework, where the policy head outputs refinement actions and the value head estimates the expected return for advantage-based learning \cite{mnih2016asynchronous,schulman2017ppo}:
\begin{equation}
\mathcal{L}_{\mathrm{RL}} =
-\mathbb{E}[\log \pi_\phi(a_k | x_k) A_k].
\label{eq:policy_gradient}
\end{equation}
Conceptually, Eq. \eqref{eq:policy_gradient} represents the advantage-weighted policy-gradient objective. In implementation, we use the PPO clipped surrogate objective with a clipping parameter of 0.20, together with the value-loss and entropy-regularization terms described above.

\subsection{Training strategy}

The conditional MeanFlow backbone is first trained on paired CT images using $\mathcal{L}_{\mathrm{base}}$ to capture the underlying reconstruction dynamics. Once the backbone has converged, the controller is subsequently trained through interaction with the transformation process, where intermediate states are sampled to learn spatially adaptive refinement policies. During this phase, the backbone remains fixed, ensuring a stable and consistent transformation model.

The controller was trained for 20 epochs with a batch size of 4 and an agent learning rate of $1 \times 10^{-5}$. At inference time, both components are integrated into a unified pipeline. The conditional MeanFlow backbone performs global reconstruction, while the controller adaptively redistributes refinement across spatial regions, resulting in the final reconstructed CT image. For implementation, all CT images were processed at a spatial resolution of $512 \times 512$ pixels. The regional controller divides each image into a $16 \times 16$ tile grid, resulting in 256 spatial tiles, with each tile corresponding to an effective image region of $32 \times 32$ pixels. During adaptive refinement, the controller can allocate up to 16 additional regional refinement steps to the selected regions.

\subsection{Experimental Data}

Experiments are carried out using paired CT images obtained from the National Lung Screening Trial (NLST) dataset, which is publicly accessible through The Cancer Imaging Archive (TCIA) \cite{nlst2013tcia,clark2013tcia}. This dataset consists of low-dose thoracic CT scans acquired for lung cancer screening. The NLST collection includes low-dose helical CT scans from participants in the CT arm of the study, with three annual screening exams acquired at baseline and during follow-up years \cite{nlstcdasimages}. To construct paired data, we restricted our selection to CT scans acquired from GE Healthcare manufacturer, to ensure consistency in the pairing procedure. Each image pair was generated from the same CT acquisition reconstructed using different GE reconstruction settings. Specifically, the paired images were constructed using GE scans with slice thicknesses of 1.25 mm and 2.5 mm and reconstruction kernels including Lung and Standard. In this study, GE Lung was treated as the standard/reference protocol, while GE Standard was treated as the non-standard/input protocol. Therefore, the paired images correspond to the same patient, same scan session, same anatomical coverage, and matched slice locations. The dataset is split at the patient level to prevent data leakage, with 17 patients used for training, 4 for validation, and 2 for testing, corresponding to 15,000, 3,000, and 1,500 DICOM slice pairs, respectively.

In addition, a separate subset of 5 patients with annotated tumor regions of interest (ROIs) is used for region-specific evaluation. These tumor annotations are obtained from the NLSTseg pixel-level lung cancer dataset, which is publicly available through Zenodo \cite{chen2025nlstseg,lin2025nlstsegzenodo}. The NLSTseg dataset is derived from NLST low-dose CT images and provides manually reviewed pixel-level annotations of lung lesions, enabling targeted evaluation within clinically relevant tumor regions \cite{chen2025nlstseg}. The inclusion of these ROIs facilitates a more detailed assessment of reconstruction performance beyond global image quality metrics.


Although the study includes 19,500 paired CT slices, corresponding to 39,000 individual CT images, the number of unique patients and scanner settings remains limited. All training, validation, and testing splits were performed strictly at the patient level to prevent data leakage between correlated slices.

\subsection{Evaluation Metrics}

Model performance was evaluated using radiomic feature consistency and image quality metrics. Radiomic features were extracted from tumor ROIs using PyRadiomics \cite{vangriethuysen2017pyradiomics}, including first-order, Gray Level Co-occurrence Matrix (GLCM), Gray Level Dependence Matrix (GLDM), Gray Level Run Length Matrix (GLRLM), Gray Level Size Zone Matrix (GLSZM), and Neighboring Gray Tone Difference Matrix (NGTDM).

\textbf{Radiomic Consistency.}
We compute the Concordance Correlation Coefficient (CCC) for each radiomic feature \cite{lin1989ccc}:
\begin{equation}
CCC(f) = \frac{2\rho_f \sigma_{s,f}\sigma_{t,f}}{\sigma_{s,f}^2 + \sigma_{t,f}^2 + (\mu_{s,f} - \mu_{t,f})^2},
\end{equation}
where $s$ and $t$ denote the reference and evaluated images, respectively. CCC is computed feature-wise and reported as class-wise and overall averages.

\textbf{Image Quality.}
We use Peak Signal-to-Noise Ratio (PSNR) and Structural Similarity Index (SSIM), defined as
\begin{align}
\mathrm{PSNR} &=
10\log_{10}\left(\frac{MAX^2}{MSE}\right),
\label{eq:psnr}\\
\mathrm{SSIM}(x,y) &=
\frac{(2\mu_x\mu_y+C_1)(2\sigma_{xy}+C_2)}
{D_{\mathrm{SSIM}}},
\label{eq:ssim}\\
D_{\mathrm{SSIM}} &=
(\mu_x^2+\mu_y^2+C_1)(\sigma_x^2+\sigma_y^2+C_2).
\end{align}
Both metrics are computed within tumor ROIs and over the full image.

\textbf{Noise Analysis.}
Noise characteristics are analyzed using the Noise Power Spectrum (NPS) computed from homogeneous regions \cite{kijewski1987nps,boedeker2007nps}.
\begin{equation}
NPS(f) = \frac{1}{N} \left| \mathcal{F}\{x - \bar{x}\} \right|^2.
\end{equation}


\section{Results}

\subsection{Radiomic Feature Consistency on Tumor ROI}

Table~\ref{tab:ccc} presents the class-wise concordance correlation coefficient (CCC) computed within tumor ROIs. The baseline results exhibit notable variability, particularly for higher-order texture features such as the Gray Level Co-occurrence Matrix (GLCM) and Neighborhood Gray Tone Difference Matrix (NGTDM), highlighting their sensitivity to differences in acquisition settings.

STAN-CT improves feature consistency across most categories, achieving CCC values of 0.88, 0.83, and 0.64 for first-order, GLCM, and GLDM features. However, its performance remains comparatively limited for GLRLM, GLSZM, and GLDM features. The conditional MeanFlow (CMF) model further enhances radiomic consistency, with a pronounced improvement in NGTDM features, achieving a CCC of 0.90 compared to 0.80 for STAN-CT. The proposed RA-CMF demonstrates the strongest overall performance across the feature classes, with consistent gains over CMF observed in texture-based features such as GLRLM and GLSZM.

The overall CCC values are summarized in Table~\ref{tab:overall_ccc}. RA-CMF attains the highest overall CCC of 0.93, outperforming RadiomicGAN (0.92), CMF (0.89), STAN-CT (0.75), and the CNN-baseline (0.71).

\begin{table*}
\centering
\captionof{table}{Class-wise concordance correlation coefficient (CCC) for radiomic features within tumor ROIs. Each column indicates the mean±std CCC values of lung tumor ROIs for a specific radiomic feature group. Higher value is better.}
\label{tab:ccc}
\small
\setlength{\tabcolsep}{4pt}
\begin{tabular*}{\textwidth}{@{\extracolsep{\fill}}lcccccc}
\toprule
\textbf{Method} & \textbf{First-order} & \textbf{GLCM} & \textbf{GLRLM} & \textbf{GLSZM} & \textbf{GLDM} & \textbf{NGTDM} \\
\midrule
Raw Input & $0.59 \pm 0.20$ & $0.45 \pm 0.27$ & $0.54 \pm 0.25$ & $0.54 \pm 0.25$ & $0.52 \pm 0.21$ & $0.40 \pm 0.26$ \\
CNN-Baseline & $0.79 \pm 0.11$ & $0.55 \pm 0.28$ & $0.79 \pm 0.14$ & $0.79 \pm 0.14$ & $0.79 \pm 0.12$ & $0.48 \pm 0.25$ \\
STAN-CT & $0.88 \pm 0.14$ & $0.83 \pm 0.20$ & $0.65 \pm 0.29$ & $0.65 \pm 0.29$ & $0.64 \pm 0.22$ & $0.80 \pm 0.25$ \\
RadiomicGAN & $0.91 \pm 0.05$ & $0.89 \pm 0.13$ & $0.91 \pm 0.06$ & $0.91 \pm 0.06$ & $0.94 \pm 0.04$ & $0.90 \pm 0.05$ \\
CMF & $\mathbf{0.97 \pm 0.05}$ & $0.91 \pm 0.14$ & $0.83 \pm 0.20$ & $0.83 \pm 0.20$ & $0.86 \pm 0.12$ & $0.90 \pm 0.09$ \\
RA-CMF (Ours) & $0.96 \pm 0.06$ & $\mathbf{0.93 \pm 0.11}$ & $\mathbf{0.91 \pm 0.09}$ & $\mathbf{0.91 \pm 0.09}$ & $\mathbf{0.94 \pm 0.05}$ & $\mathbf{0.96 \pm 0.03}$ \\
\bottomrule
\end{tabular*}
\end{table*}

\subsection{Tumor ROI Image Quality}

Table~\ref{tab:psnr_ssim} summarizes image quality evaluation using peak signal-to-noise ratio (PSNR) and structural similarity index (SSIM) within tumor ROIs. The raw input results show relatively low PSNR (17.81\,dB) and SSIM (0.53), reflecting the degradation typically observed in the input CT images. The CNN-baseline improves both metrics, achieving a PSNR of 20.29\,dB and an SSIM of 0.69. STAN-CT further improves both metrics, achieving a PSNR of 26.58\,dB and an SSIM of 0.88. RadiomicGAN achieves a PSNR of 27.30\,dB and an SSIM of 0.91. The CMF model further enhances image quality, increasing PSNR to 29.57\,dB and SSIM to 0.94. The proposed RA-CMF achieves the best performance among all methods, with a PSNR of 31.94\,dB and an SSIM of 0.97.

\begin{figure*}[!t]
\centering
\begin{subfigure}{0.99\textwidth}
    \centering
    \includegraphics[width=\linewidth,trim=0 0 0 0pt,clip]{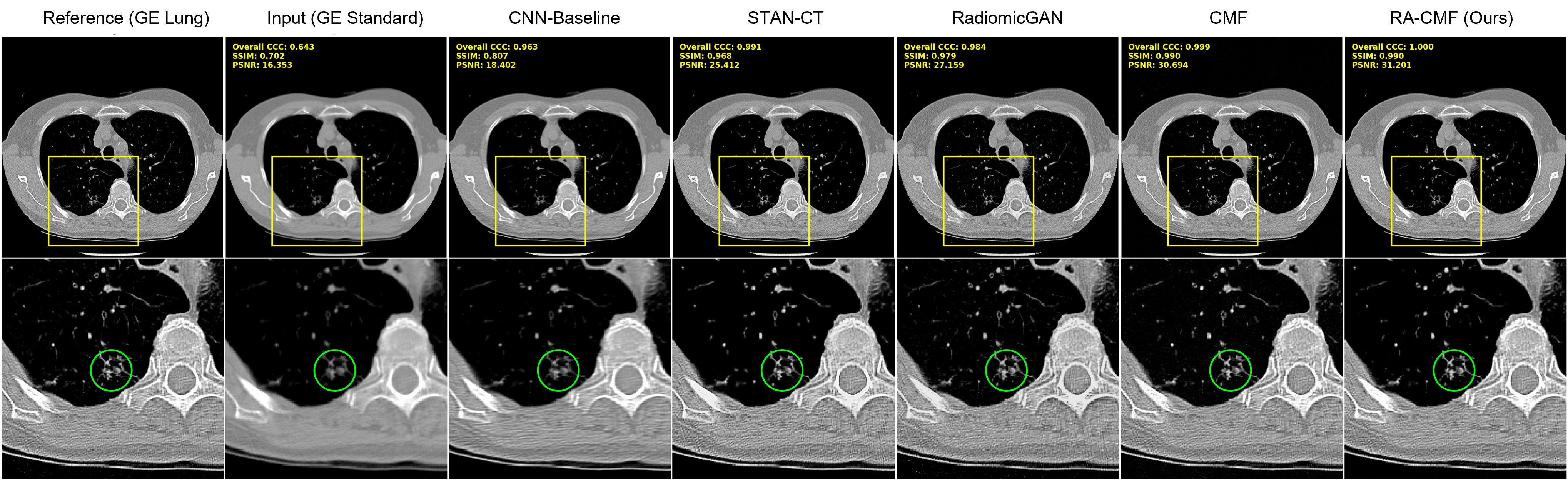}
    \caption{Subject ID 202610, slice 55. ROI metrics (CCC,SSIM,PSNR): Input (0.643, 0.702, 16.353 dB), CNN-Baseline (0.963, 0.807, 18.402 dB), STAN-CT (0.991, 0.968, 25.412 dB), RadiomicGAN (0.984, 0.979, 27.159 dB), CMF (0.999, 0.990, 30.694 dB), and RA-CMF (1.000, 0.990, 31.201 dB).
}
\end{subfigure}
\hfill

\vspace{0.5em}

\begin{subfigure}{0.99\textwidth}
    \centering
    \includegraphics[width=\linewidth,trim=0 0 0 0pt,clip]{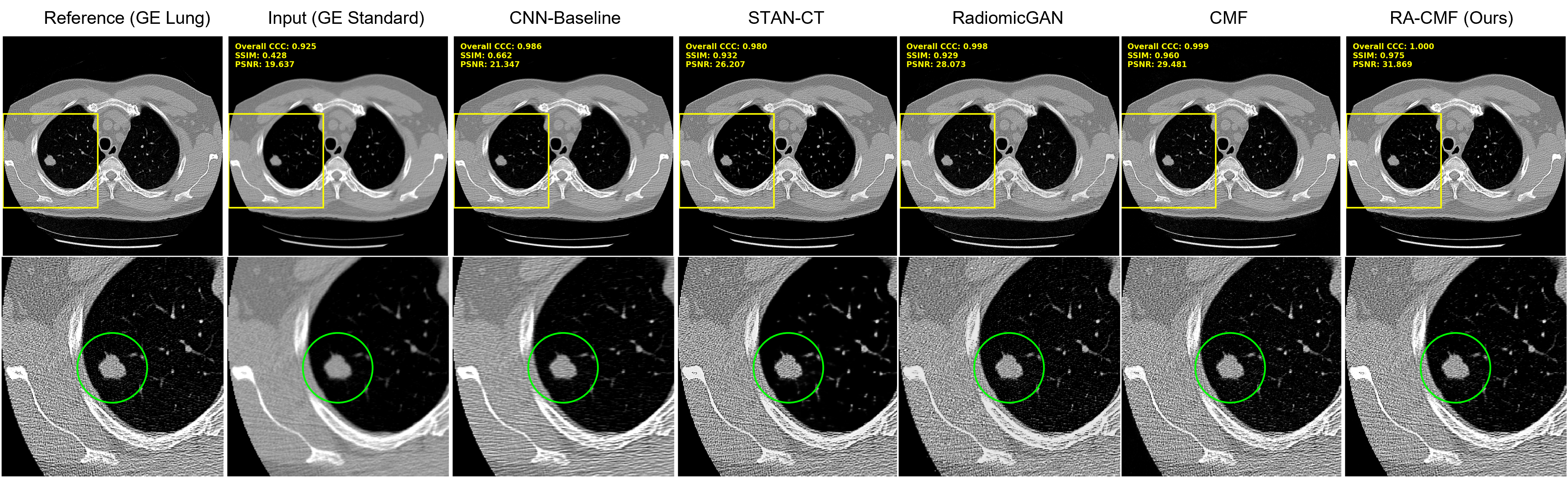}
    \caption{Subject ID 201519, slice 55. ROI metrics (CCC,SSIM,PSNR): Input (0.925, 0.428, 19.637 dB), CNN-Baseline (0.986, 0.662, 21.347 dB), STAN-CT (0.980, 0.932, 26.207 dB), RadiomicGAN (0.998, 0.929, 28.073 dB), CMF (0.999, 0.960, 29.481 dB), and RA-CMF (1.000, 0.975, 31.869 dB).
    }
\end{subfigure}
\hfill

\caption{Representative CT reconstruction results for two tumor-containing test slices: (a) subject ID 202610, slice 55, and (b) subject ID 201519, slice 55. For each case, the top row shows the full CT image with the tumor ROI indicated by a yellow box, and the bottom row shows the corresponding enlarged ROI with the tumor highlighted by a green circle. Images are displayed within the intensity window of $[-800, 600]$ HU. The values shown in each panel denote ROI-level CCC, SSIM, and PSNR (dB). RA-CMF provides the closest reconstruction to the reference across both examples, with improved quantitative agreement while preserving local tumor structure.}
\label{fig:roi_qualitative}
\end{figure*}

\begin{table*}[t]
\centering

\begin{minipage}{0.48\textwidth}
\centering
\caption{Overall CCC within tumor ROIs. Higher score is better.}
\label{tab:overall_ccc}
\begin{tabular*}{\linewidth}{@{\extracolsep{\fill}}lc}
\hline
\textbf{Method} & \textbf{Overall CCC} \\
\hline
Raw Input & 0.51 ± 0.24 \\
CNN-Baseline & 0.71 ± 0.22 \\
STAN-CT & 0.75 ± 0.25 \\
RadiomicGAN & 0.92 ± 0.08 \\
CMF & 0.89 ± 0.15 \\
RA-CMF (Ours) & \textbf{0.93 ± 0.09} \\
\hline
\end{tabular*}
\end{minipage}
\hfill
\begin{minipage}{0.49\textwidth}
\centering
\caption{PSNR and SSIM within tumor ROIs. Higher score is better.}
\label{tab:psnr_ssim}
\begin{tabular*}{\linewidth}{@{\extracolsep{\fill}}lcc}
\hline
\textbf{Method} & \textbf{PSNR (dB)} & \textbf{SSIM} \\
\hline
Raw Input & $17.81 \pm 1.84$ & $0.53 \pm 0.13$ \\
CNN-Baseline & 20.29 $\pm$ 1.89 & 0.69 $\pm$ 0.10 \\
STAN-CT & 26.58 $\pm$ 2.48 & 0.88 $\pm$ 0.16 \\
RadiomicGAN & 27.30 $\pm$ 2.79 & 0.91 $\pm$ 0.07 \\
CMF & 29.57 $\pm$ 1.98 & 0.94 $\pm$ 0.03 \\
RA-CMF (Ours) & \textbf{31.94 $\pm$ 2.64} & \textbf{0.97 $\pm$ 0.03} \\
\hline
\end{tabular*}
\end{minipage}
\end{table*}

\subsection{Overall Image Quality Evaluation}

The full-image PSNR and SSIM results are summarized in Table~\ref{tab:full_psnr_ssim}. The raw input, CNN-baseline, STAN-CT, and RadiomicGAN methods yield PSNR values of 22.59\,dB, 25.21\,dB, 30.41\,dB, and 29.35\,dB, respectively. However, RadiomicGAN shows a slightly higher SSIM of 0.92 compared with 0.90 for STAN-CT, which suggests that, although its PSNR is slightly lower, it preserves global structural similarity well.

On the other hand, the proposed CMF model exhibits a significant performance boost on these two aspects, achieving 32.09\,dB PSNR and 0.90 SSIM. This implies that the proposed model performs well on image reconstruction without sacrificing much of the anatomical structure in the reconstructed image. Further improvement can be seen using RA-CMF, which is able to generate image pairs with 34.23\,dB PSNR and 0.95 SSIM in the full test set of 1,500 slices.

\begin{table}[!ht]
\centering
\caption{Comparison on overall image PSNR and SSIM. Higher score is better.}
\label{tab:full_psnr_ssim}
\begin{tabular*}{\linewidth}{@{\extracolsep{\fill}}lcc}
\hline
\textbf{Method} & \textbf{PSNR (dB)} & \textbf{SSIM} \\
\hline
Raw Input & $22.59 \pm 2.27$ & $0.67 \pm 0.07$ \\
CNN-Baseline & $25.21 \pm 2.44$ & $0.81 \pm 0.05$ \\
STAN-CT & $30.41 \pm 1.76$ & $0.90 \pm 0.03$ \\
RadiomicGAN & $29.35 \pm 1.26$ & $0.92 \pm 0.02$ \\
CMF & $32.09 \pm 1.49$ & $0.90 \pm 0.01$ \\
RA-CMF (Ours) & $\mathbf{34.23 \pm 1.71}$ & $\mathbf{0.95 \pm 0.01}$ \\
\hline
\end{tabular*}
\end{table}

\subsection{Spatial Refinement Analysis}

\begin{figure}[t]
\centering
\begin{subfigure}{0.48\columnwidth}
\centering
\includegraphics[width=\linewidth,trim=0 0 0 40pt,clip]{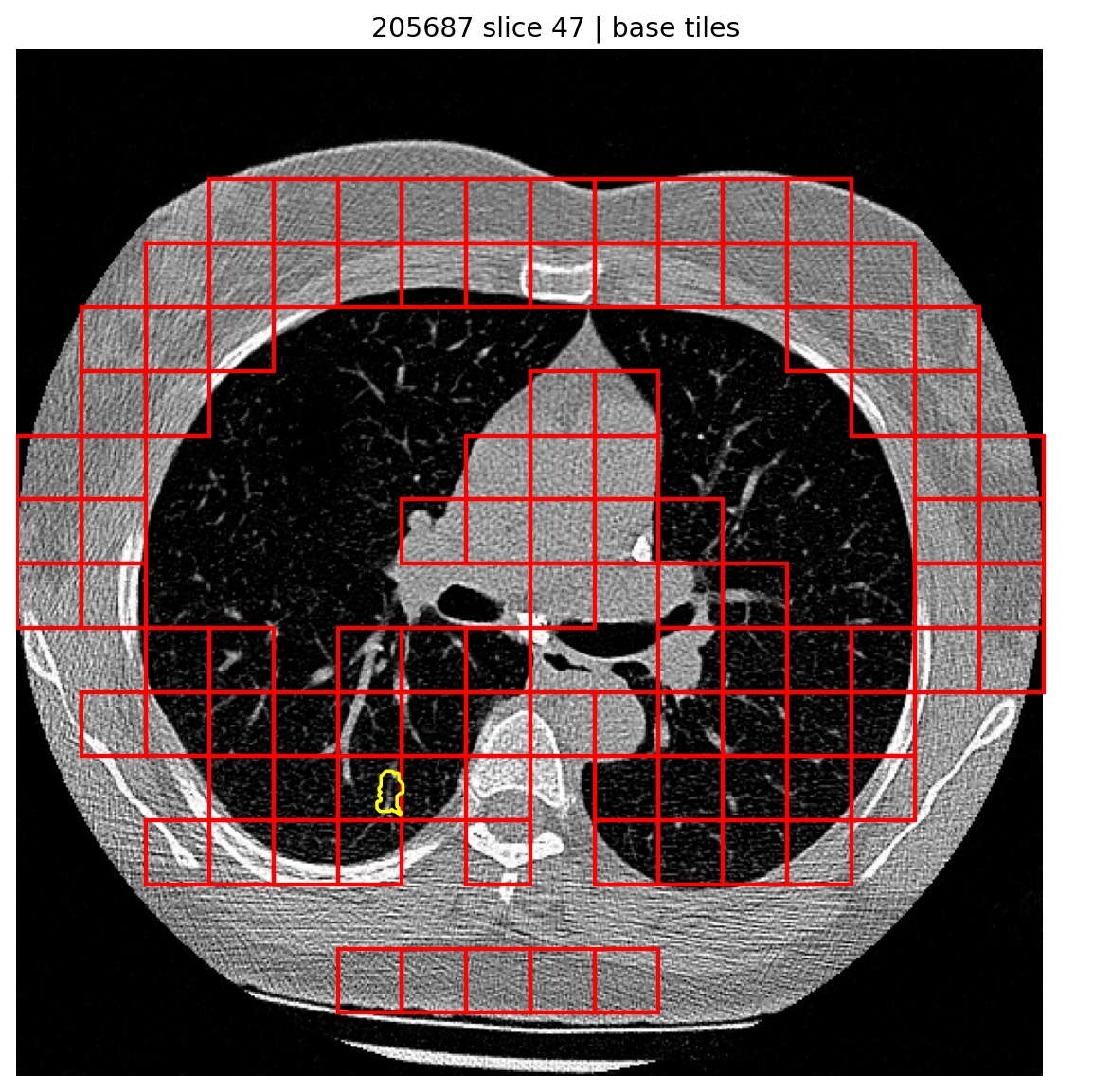}
\caption{Refinement on CMF}
\end{subfigure}
\hfill
\begin{subfigure}{0.48\columnwidth}
\centering
\includegraphics[width=\linewidth,trim=0 0 0 40pt,clip]{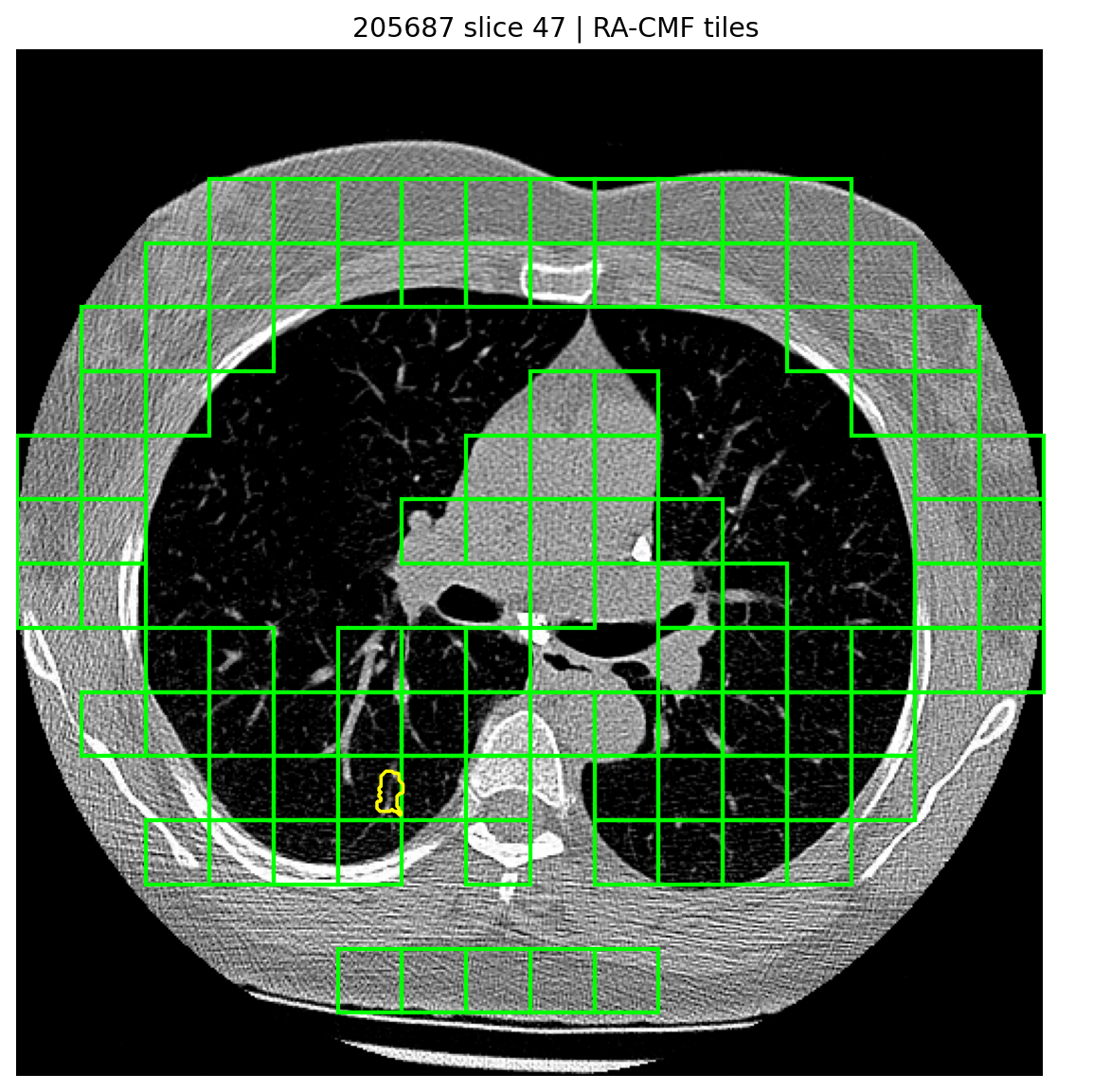}
\caption{Refined by RA-CMF}
\end{subfigure}

\caption{Visualization of regions to be refined (red tile). Notice the clustering of refinement tiles near the tumor ROI (yellow mask), implying that RA-CMF focuses refinement effort (green tile) on relevant regions (display HU window of $[-800, 600]$).}
\label{fig:tile_selection}
\end{figure}

The difference between the base tile grid and refinement regions selected by RA-CMF is shown in Figure~\ref{fig:tile_selection}. The base grid represents an initial and crude selection for image processing, whereas the selection of the refined regions depends on the actual content of the image.

However, the refinement regions do not show a uniform spatial distribution. On the contrary, the majority of them appear to cluster in more complex structural regions, namely, the lung parenchyma, with a slight concentration in the vicinity of the tumor ROIs. It is worth noting that the clustering of refinement tiles implies that this approach focuses on more complex and harder-to-reconstruct anatomical regions.

\subsection{Noise Power Spectrum Analysis}

\begin{figure*}[!t]
\centering

\begin{subfigure}{0.32\textwidth}
    \centering
    \includegraphics[width=\linewidth]{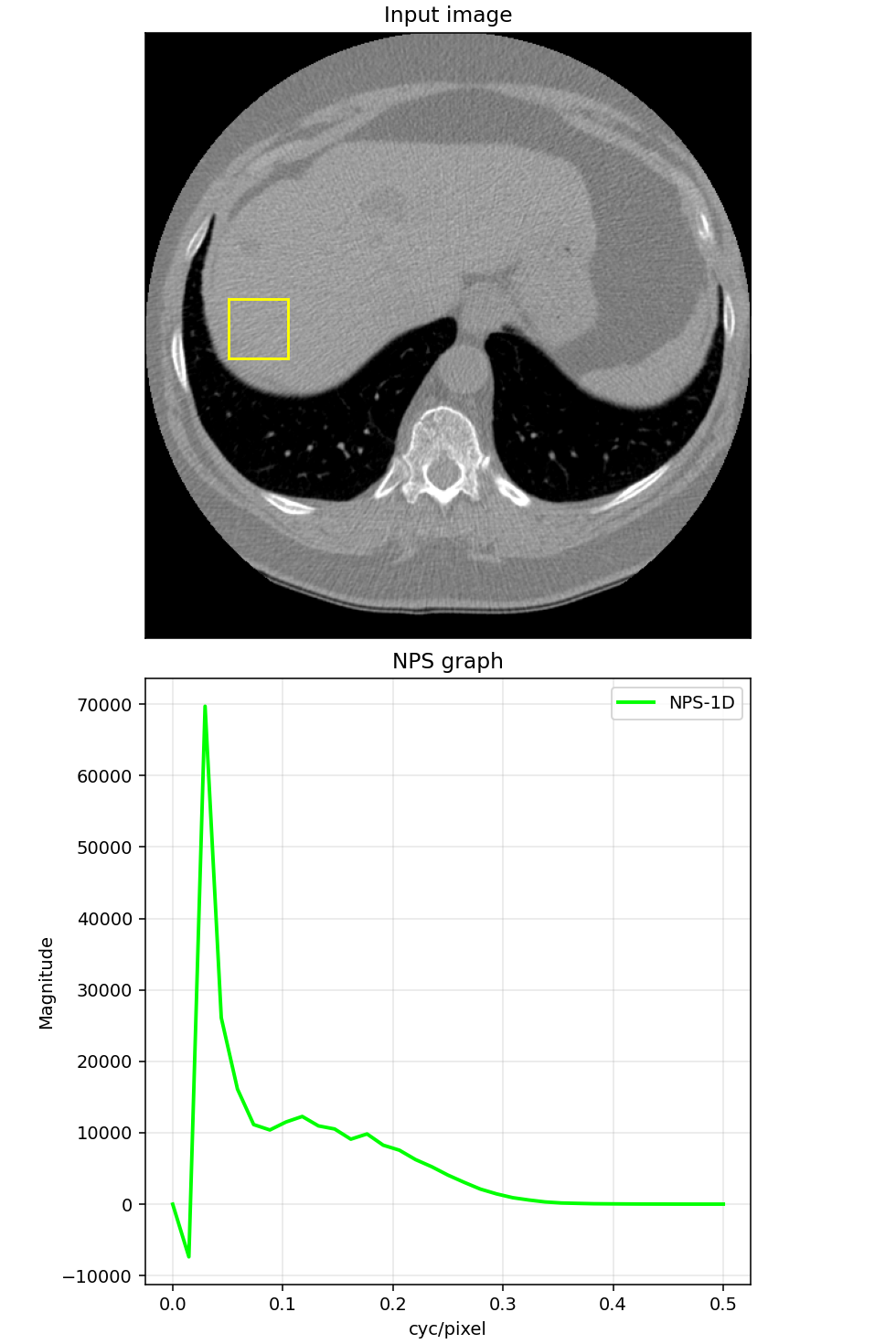}
    \caption{Input}
\end{subfigure}
\hfill
\begin{subfigure}{0.32\textwidth}
    \centering
    \includegraphics[width=\linewidth]{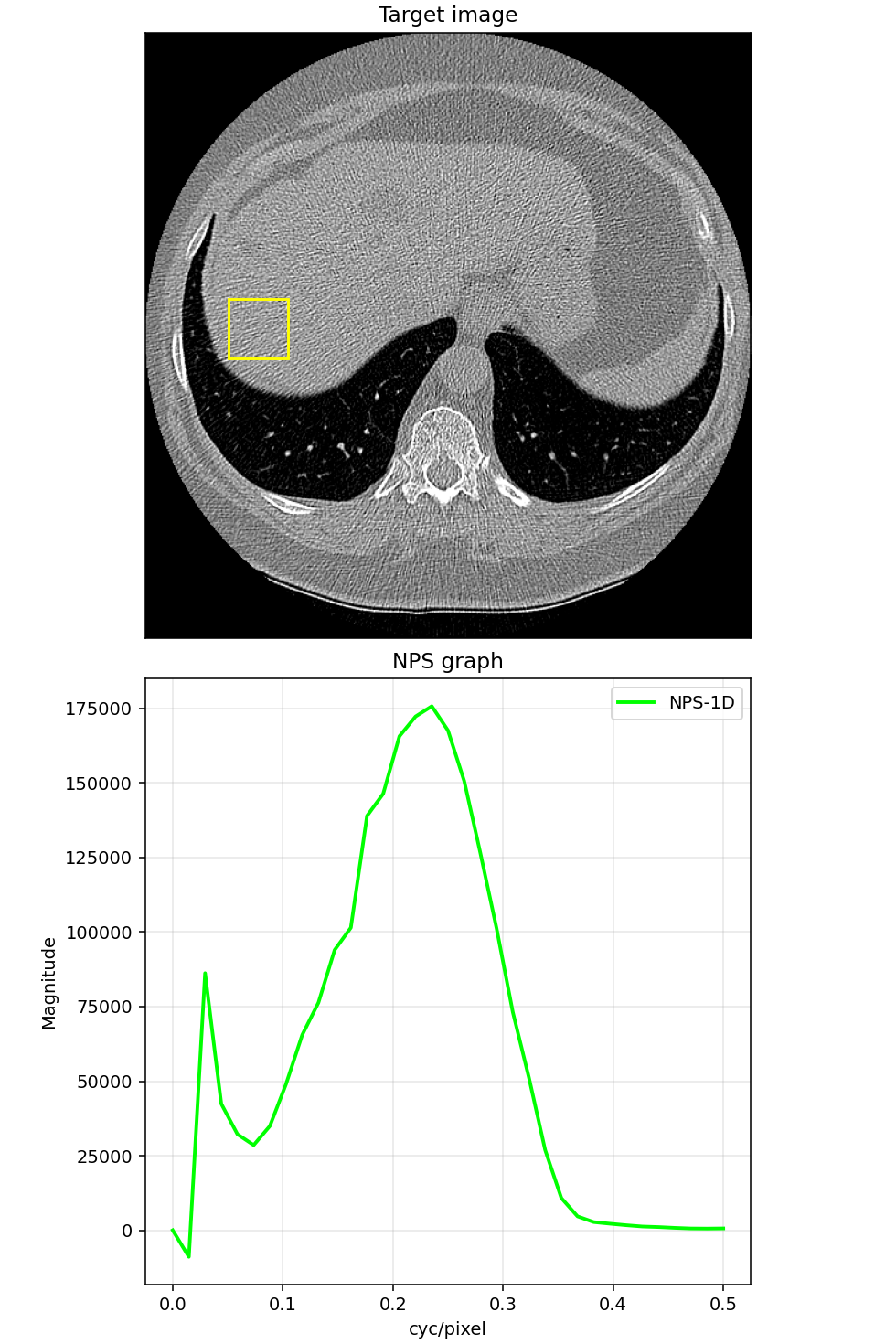}
    \caption{RA-CMF}
\end{subfigure}
\hfill
\begin{subfigure}{0.32\textwidth}
    \centering
    \includegraphics[width=\linewidth]{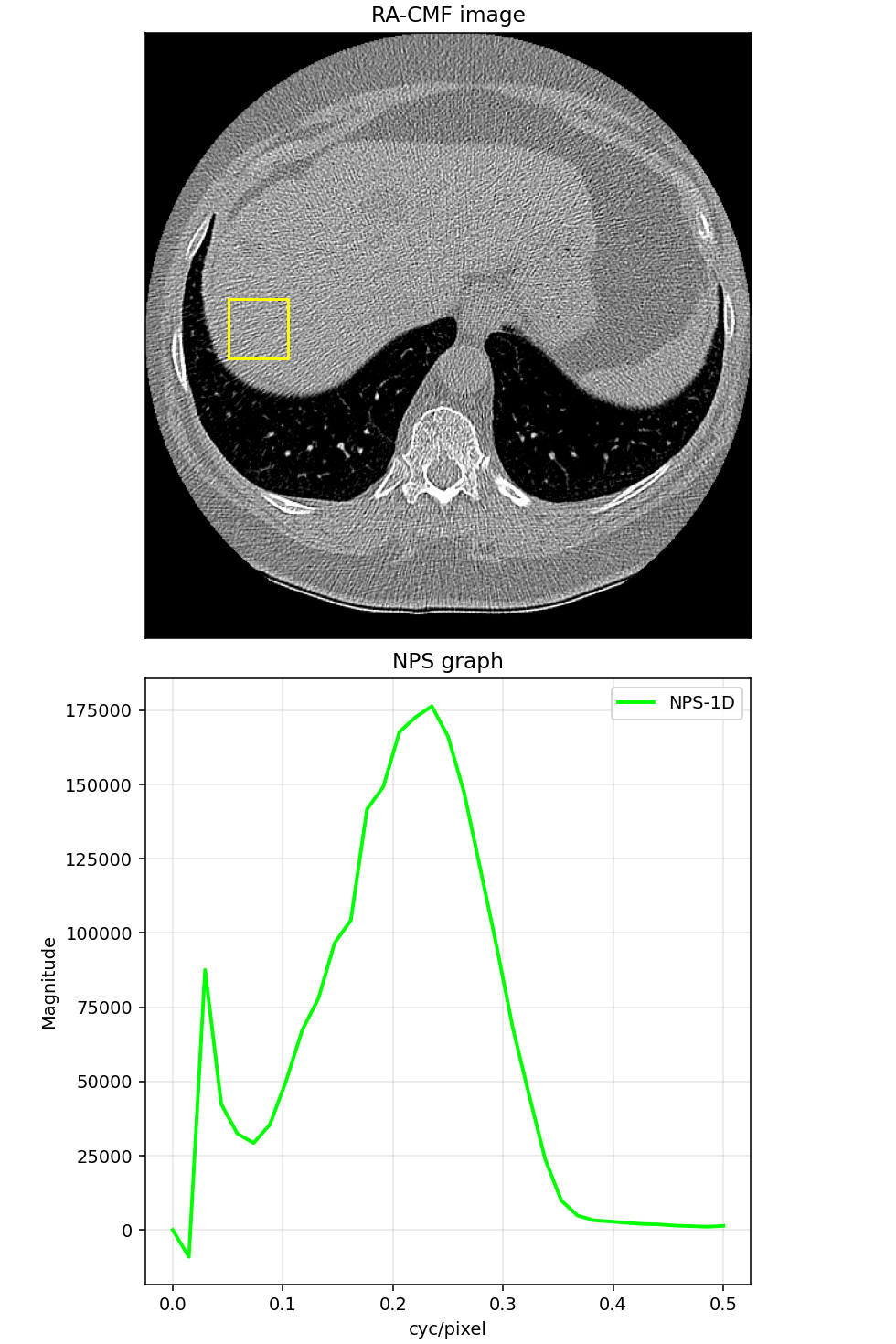}
    \caption{Target}
\end{subfigure}

\caption{Noise Power Spectrum (NPS) analysis computed from a homogeneous region (yellow box). The input image exhibits a noise–frequency distribution that differs from the target, whereas RA-CMF produces an NPS profile that more closely matches the target image, indicating improved noise characteristics and better preservation of frequency content.
}
\label{fig:nps_analysis}
\end{figure*}

To further understand the characteristics of noise, we analyze the Noise Power Spectrum (NPS), which describes how variance in noise is distributed with respect to spatial frequencies. It is a commonly used method to assess both the magnitude and noise distribution in CT imaging \cite{kijewski1987nps,boedeker2007nps}. The NPS calculation relies on a homogeneous ROI, as shown in Figure~\ref{fig:nps_analysis}. The ROI is chosen from a region (60 slices) where the underlying structures are relatively homogeneous in order to measure only the noise characteristics.

The NPS for the input image shows a pronounced peak near low spatial frequencies, indicating a structured noise pattern and inhomogeneity in the input image. However, the RA-CMF reconstruction shows reduced dominance of the low-frequency components and a power distribution that more closely resembles the target image. In particular, the RA-CMF reconstruction has a very similar NPS profile as the target image. The differences between the input image and target images are apparent.

\section{Discussion}

The first observation is that the modeling of image reconstruction as a conditional transformation results in performance improvement over baseline models and competitive or improved performance relative to GAN-based models. The second conclusion drawn from this experiment is that region-aware refinement does not simply improve regions surrounding the tumor regions, but helps to produce overall image reconstruction with better structural preservation, especially for the higher SSIM score.

The comparison between CMF and RA-CMF provides a component-level assessment of the proposed region-aware refinement module. Since both methods use the same conditional MeanFlow backbone, the improvement from CMF to RA-CMF reflects the contribution of the adaptive regional controller and local refinement process. Within tumor ROIs, RA-CMF improves PSNR from 29.57 dB to 31.94 dB and SSIM from 0.94 to 0.97 compared with CMF. Similarly, the overall CCC improves from 0.89 to 0.93. In the full-image evaluation, RA-CMF improves PSNR from 32.09 dB to 34.23 dB and SSIM from 0.90 to 0.95. These results indicate that the region-aware refinement module contributes meaningful performance improvements beyond the conditional MeanFlow backbone alone.

Contrary to the base tiling approach, where updates are applied universally, this region-aware strategy avoids updating of homogeneous anatomical regions, thus limiting computation to areas of richer structural features. This selective refinement results in improved local image quality, which is also observed in improved consistency of radiomic characteristics in tumor regions.

This analysis suggests that the proposed RA-CMF model is capable of generating realistic image reconstructions not only by controlling the magnitude but also by controlling the frequency characteristics of the target image noise pattern.

Future work will extend the evaluation to larger external cohorts, multiple scanner vendors, and more diverse acquisition protocols to further assess generalizability.

\section{Conclusion}

Radiomic feature extraction from CT images is highly sensitive to variations introduced during image acquisition, making image reconstruction and protocol standardization important for improving reliability. In this work, we propose a conditional MeanFlow-based framework for image-domain CT reconstruction, combined with a region-aware refinement strategy to better handle spatially heterogeneous degradation.

The method is evaluated on low-dose CT data from the NLST dataset, where it demonstrates consistent improvements in both radiomic feature reproducibility and overall image quality. In particular, RA-CMF achieves the highest overall CCC and consistently strong performance across the evaluated radiomic feature classes, while also showing clear improvements in PSNR and SSIM compared with the CNN-Baseline, STAN-CT, and RadiomicGAN methods.

Further analysis shows that the region-aware refinement mechanism naturally focuses on structurally complex areas, such as regions near tumor ROIs, while avoiding unnecessary updates in more homogeneous regions. Furthermore, the Noise Power Spectrum (NPS) analysis suggests that the proposed approach more faithfully preserves the noise characteristics of the target image.

In conclusion, the proposed framework offers a structured and effective approach to image-domain CT reconstruction by combining conditional transformation modeling with adaptive spatial refinement. By improving both visual quality and quantitative consistency, while maintaining clinically meaningful structures, the method demonstrates strong potential for enhancing the reliability of downstream radiomic analysis and supporting robust quantitative imaging workflows.

\section*{Acknowledgments}

\paragraph*{Funding.}
This work was supported by the Patient-Centered Outcomes Research Institute (PCORI) under Award No. ME-2024C1-33497.

\paragraph*{AI disclosure.}
During the preparation of this work, the authors used OpenAI ChatGPT and Google Gemini to assist with language refinement, organization, and improvement of clarity and readability. The authors reviewed and edited the AI-assisted content as needed and take full responsibility for the content of the manuscript.

\printbibliography

@article{siegel2024cancer,
  author  = {Rebecca L. Siegel and Angela N. Giaquinto and Ahmedin Jemal},
  title   = {Cancer statistics, 2024},
  journal = {CA: A Cancer Journal for Clinicians},
  volume  = {74},
  number  = {1},
  pages   = {12--49},
  year    = {2024},
  doi     = {10.3322/caac.21820},
  url     = {https://acsjournals.onlinelibrary.wiley.com/doi/full/10.3322/caac.21820}
}

@misc{who2026lungcancer,
  author       = {{World Health Organization}},
  title        = {Lung cancer},
  year         = {2026},
  howpublished = {\url{https://www.who.int/news-room/fact-sheets/detail/lung-cancer}},
  note         = {Accessed: 2026-04-25}
}

@article{siegel2025cancer,
  author  = {Rebecca L. Siegel and Angela N. Giaquinto and Ahmedin Jemal},
  title   = {Cancer statistics, 2025},
  journal = {CA: A Cancer Journal for Clinicians},
  volume  = {75},
  number  = {1},
  pages   = {10--45},
  year    = {2025},
  doi     = {10.3322/caac.21871},
  url     = {https://acsjournals.onlinelibrary.wiley.com/doi/10.3322/caac.21871}
}

@misc{seer2026lungbronchus,
  author       = {{National Cancer Institute}},
  title        = {Cancer Stat Facts: Lung and Bronchus Cancer},
  howpublished = {\url{https://seer.cancer.gov/statfacts/html/lungb.html}},
  year         = {2026},
  note         = {Surveillance, Epidemiology, and End Results Program. Accessed: 2026-04-25}
}

@article{aberle2011reduced,
  author  = {Denise R. Aberle and Amanda M. Adams and Christine D. Berg and William C. Black and Jonathan D. Clapp and Richard M. Fagerstrom and Ilana F. Gareen and Constantine Gatsonis and Pamela M. Marcus and JoRean D. Sicks},
  title   = {Reduced lung-cancer mortality with low-dose computed tomographic screening},
  journal = {The New England Journal of Medicine},
  volume  = {365},
  number  = {5},
  pages   = {395--409},
  year    = {2011},
  doi     = {10.1056/NEJMoa1102873},
  url     = {https://www.nejm.org/doi/full/10.1056/NEJMoa1102873}
}

@article{aerts2014decoding,
  author  = {Hugo J. W. L. Aerts and Emmanuel Rios Velazquez and Ralph T. H. Leijenaar and Chintan Parmar and Patrick Grossmann and Sara Carvalho and Johan Bussink and Ren{\'e} Monshouwer and Benjamin Haibe-Kains and Derek Rietveld and Frank Hoebers and Michelle M. Rietbergen and C. Ren{\'e} Leemans and Andre Dekker and John Quackenbush and Robert J. Gillies and Philippe Lambin},
  title   = {Decoding tumour phenotype by noninvasive imaging using a quantitative radiomics approach},
  journal = {Nature Communications},
  volume  = {5},
  pages   = {4006},
  year    = {2014},
  doi     = {10.1038/ncomms5006},
  url     = {https://doi.org/10.1038/ncomms5006}
}

@article{lambin2017radiomics,
  author  = {Philippe Lambin and Ralph T. H. Leijenaar and Timo M. Deist and Jurgen Peerlings and Evelyn E. C. de Jong and Janita van Timmeren and Sebastian Sanduleanu and Ruben T. H. M. Larue and Aniek J. G. Even and Arthur Jochems and Yvonka van Wijk and Henry Woodruff and Johan van Soest and Tim Lustberg and Erik Roelofs and Wouter van Elmpt and Andre Dekker and Felix M. Mottaghy and Joachim E. Wildberger and Sean Walsh},
  title   = {Radiomics: the bridge between medical imaging and personalized medicine},
  journal = {Nature Reviews Clinical Oncology},
  volume  = {14},
  number  = {12},
  pages   = {749--762},
  year    = {2017},
  doi     = {10.1038/nrclinonc.2017.141},
  url     = {https://pubmed.ncbi.nlm.nih.gov/28975929/}
}

@article{mackin2015measuring,
  author  = {Dennis Mackin and Xenia Fave and Lei Zhang and David Fried and Jing Yang and Brandon Taylor and Edgardo Rodriguez-Rivera and Chad Dodge and A. Kyle Jones and Laurence Court},
  title   = {Measuring computed tomography scanner variability of radiomics features},
  journal = {Investigative Radiology},
  volume  = {50},
  number  = {11},
  pages   = {757--765},
  year    = {2015},
  doi     = {10.1097/RLI.0000000000000180},
  url     = {https://pubmed.ncbi.nlm.nih.gov/26115366/}
}

@article{orlhac2018postreconstruction,
  author  = {Fanny Orlhac and Sarah Boughdad and Cathy Philippe and Hugo Stalla-Bourdillon and Christophe Nioche and Laurence Champion and Micha{\"e}l Soussan and Fr{\'e}d{\'e}rique Frouin and Vincent Frouin and Ir{\`e}ne Buvat},
  title   = {A postreconstruction harmonization method for multicenter radiomic studies in {PET}},
  journal = {Journal of Nuclear Medicine},
  volume  = {59},
  number  = {8},
  pages   = {1321--1328},
  year    = {2018},
  doi     = {10.2967/jnumed.117.199935},
  url     = {https://jnm.snmjournals.org/content/59/8/1321}
}

@article{choe2019deep,
  author  = {Jooae Choe and Sang Min Lee and Kyung-Hyun Do and Gaeun Lee and Joon Beom Lee and Sang Min Lee and Joon Beom Seo},
  title   = {Deep learning--based image conversion of {CT} reconstruction kernels improves radiomics reproducibility for pulmonary nodules or masses},
  journal = {Radiology},
  volume  = {292},
  number  = {2},
  pages   = {365--373},
  year    = {2019},
  doi     = {10.1148/radiol.2019181960},
  url     = {https://pubmed.ncbi.nlm.nih.gov/31210613/}
}

@inproceedings{selim2024diffusionct,
  author    = {Md Selim and Jie Zhang and Michael A. Brooks and Ge Wang and Jin Chen},
  title     = {{DiffusionCT}: Latent Diffusion Model for {CT} Image Standardization},
  booktitle = {AMIA Annual Symposium Proceedings},
  pages     = {624--633},
  year      = {2024},
  url       = {https://pmc.ncbi.nlm.nih.gov/articles/PMC10785850/}
}

@article{lambin2012radiomics,
  author  = {Philippe Lambin and Emmanuel Rios-Velazquez and Ralph Leijenaar and Sara Carvalho and Ruud G. P. M. van Stiphout and Patrick Granton and Catharina M. L. Zegers and Robert Gillies and Ronald Boellard and Andr{\'e} Dekker and Hugo J. W. L. Aerts},
  title   = {Radiomics: Extracting more information from medical images using advanced feature analysis},
  journal = {European Journal of Cancer},
  volume  = {48},
  number  = {4},
  pages   = {441--446},
  year    = {2012},
  doi     = {10.1016/j.ejca.2011.11.036},
  url     = {https://pubmed.ncbi.nlm.nih.gov/22257792/}
}

@article{vangriethuysen2017pyradiomics,
  author  = {Joost J. M. van Griethuysen and Andriy Fedorov and Chintan Parmar and Ahmed Hosny and Nicole Aucoin and Vivek Narayan and Regina G. H. Beets-Tan and Jean-Christophe Fillion-Robin and Steve Pieper and Hugo J. W. L. Aerts},
  title   = {Computational Radiomics System to Decode the Radiographic Phenotype},
  journal = {Cancer Research},
  volume  = {77},
  number  = {21},
  pages   = {e104--e107},
  year    = {2017},
  doi     = {10.1158/0008-5472.CAN-17-0339},
  url     = {https://aacrjournals.org/cancerres/article/77/21/e104/662617/Computational-Radiomics-System-to-Decode-the}
}

@article{zwanenburg2020ibsi,
  author  = {Alex Zwanenburg and Martin Valli{\`e}res and Mahmoud A. Abdalah and Hugo J. W. L. Aerts and Vincent Andrearczyk and Apte Aditya and Ashrafinia Saeed and Spyridon Bakas and Roelof J. Beukinga and Ronald Boellaard and others},
  title   = {The Image Biomarker Standardization Initiative: Standardized Quantitative Radiomics for High-Throughput Image-based Phenotyping},
  journal = {Radiology},
  volume  = {295},
  number  = {2},
  pages   = {328--338},
  year    = {2020},
  doi     = {10.1148/radiol.2020191145},
  url     = {https://pubmed.ncbi.nlm.nih.gov/32154773/}
}

@article{haralick1973textural,
  author  = {Robert M. Haralick and K. Shanmugam and Its'hak Dinstein},
  title   = {Textural Features for Image Classification},
  journal = {IEEE Transactions on Systems, Man, and Cybernetics},
  volume  = {SMC-3},
  number  = {6},
  pages   = {610--621},
  year    = {1973},
  doi     = {10.1109/TSMC.1973.4309314},
  url     = {https://doi.org/10.1109/TSMC.1973.4309314}
}

@article{sun1983neighboring,
  author  = {Chengjun Sun and William G. Wee},
  title   = {Neighboring Gray Level Dependence Matrix for Texture Classification},
  journal = {Computer Vision, Graphics, and Image Processing},
  volume  = {23},
  number  = {3},
  pages   = {341--352},
  year    = {1983},
  doi     = {10.1016/0734-189X(83)90032-4},
  url     = {https://www.sciencedirect.com/science/article/pii/0734189X83900324}
}

@article{galloway1975texture,
  author  = {Mary M. Galloway},
  title   = {Texture Analysis Using Gray Level Run Lengths},
  journal = {Computer Graphics and Image Processing},
  volume  = {4},
  number  = {2},
  pages   = {172--179},
  year    = {1975},
  doi     = {10.1016/S0146-664X(75)80008-6},
  url     = {https://doi.org/10.1016/S0146-664X(75)80008-6}
}

@inproceedings{thibault2009texture,
  author    = {Guillaume Thibault and Bernard Fertil and Claire Navarro and Sandrine Pereira and Pierre Cau and Nicolas L{\'e}vy and Jean Sequeira and Jean-Luc Mari},
  title     = {Texture Indexes and Gray Level Size Zone Matrix: Application to Cell Nuclei Classification},
  booktitle = {Proceedings of the 10th International Conference on Pattern Recognition and Information Processing},
  year      = {2009}
}

@article{amadasun1989textural,
  author  = {M. Amadasun and R. King},
  title   = {Textural Features Corresponding to Textural Properties},
  journal = {IEEE Transactions on Systems, Man, and Cybernetics},
  volume  = {19},
  number  = {5},
  pages   = {1264--1274},
  year    = {1989},
  doi     = {10.1109/21.44046},
  url     = {https://doi.org/10.1109/21.44046}
}

@article{shafiq2018voxel,
  author  = {Muhammad Shafiq-ul-Hassan and Kujtim Latifi and Geoffrey Zhang and Ghanim Ullah and Robert Gillies and Eduardo Moros},
  title   = {Voxel Size and Gray Level Normalization of {CT} Radiomic Features in Lung Cancer},
  journal = {Scientific Reports},
  volume  = {8},
  pages   = {10545},
  year    = {2018},
  doi     = {10.1038/s41598-018-28895-9},
  url     = {https://www.nature.com/articles/s41598-018-28895-9}
}

@article{orlhac2022combat,
  author  = {Fanny Orlhac and Jakoba J. Eertink and Anne-S{\'e}gol{\`e}ne Cottereau and Jos{\'e}e M. Zijlstra and Catherine Thieblemont and Michel Meignan and Ronald Boellaard and Ir{\`e}ne Buvat},
  title   = {A Guide to {ComBat} Harmonization of Imaging Biomarkers in Multicenter Studies},
  journal = {Journal of Nuclear Medicine},
  volume  = {63},
  number  = {2},
  pages   = {172--179},
  year    = {2022},
  doi     = {10.2967/jnumed.121.262464},
  url     = {https://jnm.snmjournals.org/content/63/2/172}
}

@article{selim2020stanct,
  author  = {Md Selim and Jie Zhang and Baowei Fei and Guo-Qiang Zhang and Jin Chen},
  title   = {{STAN-CT}: Standardizing {CT} Image using Generative Adversarial Networks},
  journal = {AMIA Annual Symposium Proceedings},
  volume  = {2020},
  pages   = {1100--1109},
  year    = {2020},
  url     = {https://pubmed.ncbi.nlm.nih.gov/33936486/}
}

@inproceedings{selim2021radiomicgan,
  author    = {Md Selim and Jie Zhang and Baowei Fei and Guo-Qiang Zhang and Jin Chen},
  title     = {{CT} Image Harmonization for Enhancing Radiomics Studies},
  booktitle = {2021 IEEE International Conference on Bioinformatics and Biomedicine},
  pages     = {1057--1062},
  year      = {2021},
  doi       = {10.1109/BIBM52615.2021.9669766},
  url       = {https://arxiv.org/abs/2107.01337}
}

@inproceedings{arjovsky2017wasserstein,
  author    = {Martin Arjovsky and Soumith Chintala and L{\'e}on Bottou},
  title     = {Wasserstein Generative Adversarial Networks},
  booktitle = {Proceedings of the 34th International Conference on Machine Learning},
  series    = {Proceedings of Machine Learning Research},
  volume    = {70},
  pages     = {214--223},
  year      = {2017},
  publisher = {PMLR},
  url       = {https://proceedings.mlr.press/v70/arjovsky17a.html}
}

@inproceedings{ho2020ddpm,
  author    = {Jonathan Ho and Ajay Jain and Pieter Abbeel},
  title     = {Denoising Diffusion Probabilistic Models},
  booktitle = {Advances in Neural Information Processing Systems},
  volume    = {33},
  pages     = {6840--6851},
  year      = {2020},
  url       = {https://proceedings.neurips.cc/paper/2020/hash/4c5bcfec8584af0d967f1ab10179ca4b-Abstract.html}
}

@inproceedings{rombach2022latent,
  author    = {Robin Rombach and Andreas Blattmann and Dominik Lorenz and Patrick Esser and Bj{\"o}rn Ommer},
  title     = {High-Resolution Image Synthesis with Latent Diffusion Models},
  booktitle = {Proceedings of the IEEE/CVF Conference on Computer Vision and Pattern Recognition},
  pages     = {10684--10695},
  year      = {2022},
  url       = {https://openaccess.thecvf.com/content/CVPR2022/html/Rombach_High-Resolution_Image_Synthesis_With_Latent_Diffusion_Models_CVPR_2022_paper.html}
}

@article{geng2025meanflows,
  author  = {Zhengyang Geng and Mingyang Deng and Xingjian Bai and J. Zico Kolter and Kaiming He},
  title   = {Mean Flows for One-step Generative Modeling},
  journal = {arXiv preprint arXiv:2505.13447},
  year    = {2025},
  doi     = {10.48550/arXiv.2505.13447},
  url     = {https://arxiv.org/abs/2505.13447}
}

@book{sutton2018reinforcement,
  author    = {Richard S. Sutton and Andrew G. Barto},
  title     = {Reinforcement Learning: An Introduction},
  edition   = {2},
  publisher = {MIT Press},
  year      = {2018},
  url       = {http://incompleteideas.net/book/the-book-2nd.html}
}

@inproceedings{mnih2016asynchronous,
  author    = {Volodymyr Mnih and Adri{\`a} Puigdom{\`e}nech Badia and Mehdi Mirza and Alex Graves and Timothy Lillicrap and Tim Harley and David Silver and Koray Kavukcuoglu},
  title     = {Asynchronous Methods for Deep Reinforcement Learning},
  booktitle = {Proceedings of the 33rd International Conference on Machine Learning},
  series    = {Proceedings of Machine Learning Research},
  volume    = {48},
  pages     = {1928--1937},
  year      = {2016},
  publisher = {PMLR},
  url       = {https://proceedings.mlr.press/v48/mniha16.html}
}

@article{schulman2017ppo,
  author  = {John Schulman and Filip Wolski and Prafulla Dhariwal and Alec Radford and Oleg Klimov},
  title   = {Proximal Policy Optimization Algorithms},
  journal = {arXiv preprint arXiv:1707.06347},
  year    = {2017},
  doi     = {10.48550/arXiv.1707.06347},
  url     = {https://arxiv.org/abs/1707.06347}
}

@article{wang2004ssim,
  author  = {Zhou Wang and Alan C. Bovik and Hamid R. Sheikh and Eero P. Simoncelli},
  title   = {Image Quality Assessment: From Error Visibility to Structural Similarity},
  journal = {IEEE Transactions on Image Processing},
  volume  = {13},
  number  = {4},
  pages   = {600--612},
  year    = {2004},
  doi     = {10.1109/TIP.2003.819861},
  url     = {https://ieeexplore.ieee.org/document/1284395}
}

@article{huynhthu2008psnr,
  author  = {Quan Huynh-Thu and Mohammed Ghanbari},
  title   = {Scope of Validity of {PSNR} in Image/Video Quality Assessment},
  journal = {Electronics Letters},
  volume  = {44},
  number  = {13},
  pages   = {800--801},
  year    = {2008},
  doi     = {10.1049/el:20080522},
  url     = {https://digital-library.theiet.org/doi/10.1049/el:20080522}
}

@techreport{nayar1989shape,
  author      = {Shree K. Nayar and Yasuo Nakagawa},
  title       = {Shape from Focus},
  institution = {Columbia University},
  year        = {1989},
  url         = {https://cave.cs.columbia.edu/Statics/publications/pdfs/Nayar_TR89.pdf}
}

@misc{nlst2013tcia,
  author       = {{National Lung Screening Trial Research Team}},
  title        = {Data from the National Lung Screening Trial ({NLST})},
  year         = {2013},
  howpublished = {The Cancer Imaging Archive},
  doi          = {10.7937/TCIA.HMQ8-J677},
  url          = {https://doi.org/10.7937/TCIA.HMQ8-J677}
}

@article{clark2013tcia,
  author  = {Kenneth Clark and Bruce Vendt and Kirk Smith and John Freymann and Justin Kirby and Paul Koppel and Stephen Moore and Stanley Phillips and David Maffitt and Michael Pringle and Lawrence Tarbox and Fred Prior},
  title   = {The Cancer Imaging Archive ({TCIA}): Maintaining and Operating a Public Information Repository},
  journal = {Journal of Digital Imaging},
  volume  = {26},
  number  = {6},
  pages   = {1045--1057},
  year    = {2013},
  doi     = {10.1007/s10278-013-9622-7},
  url     = {https://doi.org/10.1007/s10278-013-9622-7}
}

@misc{nlstcdasimages,
  author       = {{National Cancer Institute}},
  title        = {{NLST} {CT} Images},
  howpublished = {Cancer Data Access System},
  year         = {2026},
  note         = {Accessed: 2026-04-25},
  url          = {https://cdas.cancer.gov/learn/nlst/images/}
}

@article{chen2025nlstseg,
  author  = {Kang-Hua Chen and others},
  title   = {{NLSTseg}: A Pixel-level Lung Cancer Dataset Based on {NLST} {LDCT} Images},
  journal = {Scientific Data},
  year    = {2025},
  doi     = {10.1038/s41597-025-05742-x},
  url     = {https://www.nature.com/articles/s41597-025-05742-x}
}

@misc{lin2025nlstsegzenodo,
  author       = {Yihui Lin},
  title        = {{NLSTseg}: A Pixel-level Lung Cancer Dataset Based on {NLST} {LDCT} Images},
  year         = {2025},
  publisher    = {Zenodo},
  version      = {v3},
  doi          = {10.5281/zenodo.14838349},
  url          = {https://zenodo.org/records/14838349}
}

@article{lin1989ccc,
  author  = {Lawrence I-Kuei Lin},
  title   = {A Concordance Correlation Coefficient to Evaluate Reproducibility},
  journal = {Biometrics},
  volume  = {45},
  number  = {1},
  pages   = {255--268},
  year    = {1989},
  doi     = {10.2307/2532051},
  url     = {https://pubmed.ncbi.nlm.nih.gov/2720055/}
}

@article{kijewski1987nps,
  author  = {Mary F. Kijewski and Philip F. Judy},
  title   = {The Noise Power Spectrum of {CT} Images},
  journal = {Physics in Medicine and Biology},
  volume  = {32},
  number  = {5},
  pages   = {565--575},
  year    = {1987},
  doi     = {10.1088/0031-9155/32/5/003},
  url     = {https://doi.org/10.1088/0031-9155/32/5/003}
}

@article{boedeker2007nps,
  author  = {Kristin L. Boedeker and Vincent N. Cooper and Michael F. McNitt-Gray},
  title   = {Application of the Noise Power Spectrum in Modern Diagnostic {MDCT}: Part I. Measurement of Noise Power Spectra and Noise Equivalent Quanta},
  journal = {Physics in Medicine and Biology},
  volume  = {52},
  number  = {14},
  pages   = {4027--4046},
  year    = {2007},
  doi     = {10.1088/0031-9155/52/14/002},
  url     = {https://doi.org/10.1088/0031-9155/52/14/002}
}

\end{document}